\documentclass{article}

\PassOptionsToPackage{numbers, compress}{natbib}
\usepackage[preprint]{neurips_2026}

\usepackage[utf8]{inputenc} 
\usepackage[T1]{fontenc}    
\usepackage{hyperref}       
\usepackage{url}            
\usepackage{booktabs}       
\usepackage{amsfonts}       
\usepackage{nicefrac}       
\usepackage{microtype}      
\usepackage[table]{xcolor} 
\usepackage{multicol}
\usepackage{multirow}
\usepackage{graphicx}    
\usepackage{subfigure}
\usepackage{enumitem}
\usepackage{amsmath} 
\usepackage{wrapfig}
\title{Learning from Reliable Latent Prompts for Visual Recognition with Missing Modalities}

\author{Taixi Chen \qquad  Nancy Guo\thanks{Corresponding author: nguo1@binghamton.edu} \vspace{.5em}\\
School of Computing \\
State University of New York at Binghamton\\
}

\begin{document}

\maketitle

\begin{abstract}
Large-scale multimodal models (LMMs) have achieved superior performance in visual recognition by synergizing information across diverse, massive-scale paired modalities. In real-world scenarios, however, missing-modality inputs are ubiquitous, causing models optimized for modality-complete data to exhibit precipitous performance degradation. Existing research has introduced prompt learning to mitigate this issue, typically by generating dynamic prompts from instance-level features, regardless of whether the input modalities are complete or partially absent. However, such input-conditioned strategies are hindered by the escalating unreliability of instance-level features; as higher missing rates increase the proportion of incomplete modalities, the resulting instability in prompt learning limits the model's performance. To address this limitation, we hypothesize that learnable latent prompts themselves encapsulate stable, modality-intrinsic priors that are decoupled from corrupted inputs. Consequently, we propose a novel paradigm: Learning from Reliable Latent Prompts. Unlike prior methods, we model input-agnostic learnable prompts as stable latent anchors that enable robust guidance and effective cross-modal knowledge compensation, even under extreme missing rates (e.g., 90\%).  Empirical results across three benchmark datasets demonstrate that our "learn-from-latent-prompts" approach achieves state-of-the-art performance across a wide range of missing-modality scenarios. Extensive experiments further confirm the effectiveness of this paradigm in providing a robust solution to the missing-modality problem.
\end{abstract}


\section{Introduction}
The essence of human intelligence is rooted in the synergistic integration of cross-modal information to achieve robust perception, a capability that modern Large-scale Multimodal Models (LMMs) aspire to emulate. By leveraging advanced pre-trained architectures~\cite{attention, mamba, vivit, chen2025tyrppg} and massive-scale aligned datasets~\cite{food-101, hatefulmemes}, these models~\cite{clip, flamingo, blip2} have demonstrated remarkable efficacy across a diverse spectrum of downstream tasks, ranging from cross-modal retrieval~\cite{clip, crossretreive, vilt} to image captioning~\cite{videocaption1, videocaption} and visual question answering~\cite{vqa, drivelm}. Despite these strides, the practical utility of LMMs is often bottlenecked by a rigid dependency on modality-complete inputs, an assumption that frequently falters in real-world deployments. Factors such as privacy constraints, sensor failures, and data collection environments often lead to partially missing modalities~\cite{first_prompt, dcp}. Such modality-incomplete inputs can significantly degrade the cross-modal alignment, leading to a precipitous degradation in performance~\cite{notrobust}. Developing robust multimodal learning frameworks that can operate reliably under varying missing-modality scenarios has therefore become a critical challenge.

Existing approaches to missing-modality learning primarily rely on cross-modal generation or joint representation learning~\cite{representation-based, notrobust, sharedfeatures}. While effective in controlled settings, these methods often introduce additional semantic noise through imperfect modality reconstruction and incur substantial computational overhead due to auxiliary model optimization~\cite{surveyformissing}. Recently, prompt-based learning has emerged as an efficient alternative, adapting pre-trained LMMs by optimizing a small set of learnable tokens while keeping the backbone frozen. Early methods employ missing-aware prompts~\cite{first_prompt}, whereas subsequent works adopt sample-aware dynamic prompting strategies that generate prompts conditioned on instance-level features~\cite{dcp}.

\begin{figure}[tb]
    \centering
    \includegraphics[width=\linewidth]{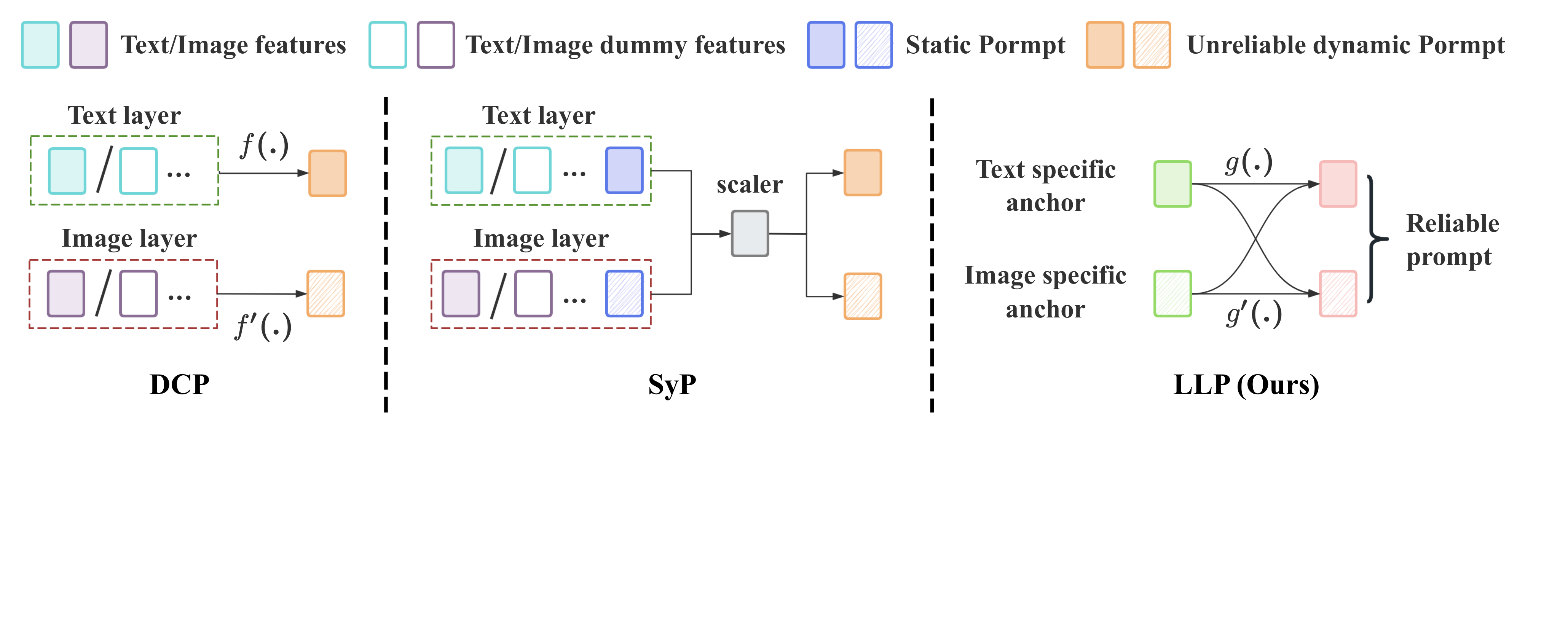}
    \vspace{-1.9cm}
    \caption{Comparison between input-conditioned prompting methods (e.g., DCP~\cite{dcp} and SyP~\cite{syp}) and our proposed LLP paradigm. Existing approaches generate prompts conditioned on instance-level features, which become unreliable under missing-modality settings due to their dependence on incomplete and modality-biased inputs. In contrast, LLP generates prompts from input-agnostic latent anchors, enabling more reliable prompt formation and robust cross-modal reasoning.}
    \label{fig:placeholder}
\end{figure}

Despite their empirical success, existing input-conditioned prompting methods~\cite{dcp, syp} fundamentally rely on incomplete observations. As a result, the learned prompts become biased estimators of full-modality semantics under high missing rates. When a large fraction of modalities is absent, the conditioning features are incomplete and modality-biased, causing the generated dynamic prompts to be dominated by unreliable or unimodal-biased signals. This suggests that directly conditioning prompt learning on incomplete inputs is intrinsically unreliable. This limitation is further supported by both the visualization and our theoretical analysis (see Appendix~\ref{appendix}), where input-conditioned methods yield representations that are locally separable within modality-specific clusters but fail to maintain a consistent class-level structure. This can be attributed to dynamic prompts collapsing to unimodal observations under high missing rates, thereby disrupting cross-modal alignment.


In this paper, we explore an alternative paradigm that decouples prompt learning from input-specific observations. We posit that learnable latent prompts, optimized independently of incomplete inputs, can capture stable modality-intrinsic priors and provide a more reliable foundation for multimodal reasoning under missingness. Based on this insight, we propose \textbf{Learning from Reliable Latent Prompts (LLP)}, a prompting framework that adopts a latent prior-driven paradigm, decoupling prompt learning from unreliable inputs. LLP introduces modality-specific latent anchors, implemented as input-agnostic learnable prompts. Unlike dynamic prompts that are directly conditioned on input features, these latent anchors are learned as input-agnostic parameters that encode modality-level priors. Crucially, these anchors serve as semantic foundations for learning refined latent prompts, rather than static offsets. To this end, we design a dual-path cross-attention mechanism that enables anchors to exchange complementary information across modalities while generating dual anchor-induced prompts (DAIP) through their interaction. Through this anchor-based interaction, LLP enables the model to approximate missing semantic information from a stable latent space, thereby mitigating the impact of input corruption even under extreme missing rates. Extensive evaluations across three widely-used datasets show that LLP consistently outperforms state-of-the-art methods, especially under high missing rates. Ablation studies further confirm the efficacy of our latent anchor design and the paradigm's robustness to diverse missing patterns. Our contributions are summarized:

\begin{itemize}
\item We propose LLP, a novel latent prior-driven prompting paradigm that decouples prompt learning from unreliable inputs and captures stable modality-level priors, leading to improved robustness under incomplete multimodal observations.

\item Specifically, LLP introduces modality-specific latent anchors as input-agnostic learnable prompts, along with a dual-path cross-attention mechanism to generate dual anchor-induced prompts, enabling input-independent prompt learning and stable cross-modal interaction, thereby achieving robust semantic compensation under severe modality missing.

\item We demonstrate state-of-the-art performance across multiple benchmarks and provide comprehensive analyses validating the robustness and generalization of the proposed approach, establishing a promising paradigm for robust multimodal learning under missing modalities.
\end{itemize}

\section{Related Works}

\label{sec:RW}
\subsection{Missing-Modality in Multi-modal Learning} Despite the remarkable strides made in multimodal learning~\cite{multisurvey, grounded, mmrl, llm2clip, clip, continual, chen2025uam}, ensuring the robustness of models under scenarios with missing modalities remains a formidable challenge~\cite{multisurvey, surveyformissing, first_prompt}. In real-world scenarios, multimodal data is often incomplete due to privacy constraints, sensor failures, and heterogeneous data collection environments~\cite{dcp, surveyformissing}. Such missing modalities introduce significant distribution shifts and disrupt cross-modal alignment, making it difficult for models to learn consistent and informative multimodal representations~\cite{surveyformissing}.To address these challenges, existing literature can be broadly categorized into three research directions: cross-modal generation~\cite{crossgen2, crossgen1, crossgen3}, joint representation learning~\cite{notrobust, representation-based, sharedfeatures}, and prompt learning~\cite{first_prompt, context-based, dcp, syp}. Cross-modal generation typically aims to reconstruct missing modalities from available ones~\cite{crossgen1, crossgen2}. Recent advances also explore leveraging auxiliary information from complete samples to guide the recovery process~\cite{crossgen3}. However, such methods are often constrained by the inherent heterogeneity across modalities and the heavy computational overhead. Joint representation learning methods~\cite{representation-based, sharedfeatures} proposed to align modalities into a unified space, but often heavily rely on data integrity. Prompt learning has recently surfaced as a promising alternative~\cite{first_prompt, syp}, requiring only the optimization of compact learnable prompts while maintaining a frozen backbone. Existing prompting methods mostly rely on the instance-level dynamic prompts~\cite{dcp, context-based, syp}. However, such dynamic prompts rely on instance-level features that may be incomplete or corrupted under missing-modality settings, which are unstable and susceptible to noise contamination. To address this issue, we propose LLP, a latent prior-driven prompting framework that decouples prompt learning from unreliable inputs and leverages stable modality-level priors.

\subsection{Prompt Learning} Prompt learning has emerged as an effective training paradigm across both natural language processing~\cite{promptlearning1, promptlearning2, promptlearning3, promptlearning4} and multimodal learning~\cite{coop, cocop, maple, dept, yang2026prompt}, demonstrating remarkable success in efficient model adaptation. CoOp~\cite{coop} first introduced learnable soft prompts for vision-language models, replacing hand-crafted templates with optimized continuous vectors to enhance task-specific adaptation. MaPLe~\cite{maple} introduces a multi-modal prompting strategy to achieve synergistic learning across both vision and language branches. Meanwhile, DePT~\cite{dept} explores a decoupled feature space to enhance the generalization and efficiency of prompt tuning. To address modality incompleteness, MMP~\cite{first_prompt} pioneered the use of missing-aware prompts for efficient adaptation. Building upon this, subsequent works such as DCP~\cite{dcp} and SyP~\cite{syp} further develop different dynamic prompting paradigms, where prompts are adaptively generated conditioned on instance-level features to capture richer semantics. These methods have demonstrated strong effectiveness in adapting large-scale vision-language models to downstream tasks, and improving robustness under missing-modality scenarios. In this paper, we propose a latent prior-driven prompting framework that decouples prompt learning from unreliable inputs by leveraging latent anchors as learnable prompts, thereby mitigating the limitations of input-conditioned prompt generation and improving robustness under missing-modality scenarios.

\section{Methods}

\subsection{Problem Formulation} We begin by providing a concise formulation of the missing-modality scenario considered in this paper. For simplicity and without loss of generality, this paper considers a multimodal dataset comprising M = 2 modalities (i.e., text and image) $m1$ and $m2$. Given a multimodal dataset $D = \{D^c, D^{m1}, D^{m2}\}$, $D^c = \{x_i^{m1}, x_i^{m2}, y_i\}$ is denoted as the modality complte subset, $D^{m1} = \{x_i^{m1}, y_i\}$ and $D^{m2} = \{x_i^{m2}, y_i\}$ are modality incomplete subsets. Similar to previous works \cite{first_prompt, dcp}, we also use dummy inputs $\tilde{x}^{m1}$ and  $\tilde{x}^{m2}$ (i.e., empty string and pixel for missing texts and images) to preserve the format of multimodal inputs, so $D^{m1}$ and $D^{m2}$ can be rewritten as $\tilde{D}^{m1} = \{x_i^{m1}, \tilde{x}^{m2}, y_i\}$ and $\tilde{D}^{m2} = \{\tilde{x}^{m1}, x_i^{m2}, y_i\}$. Thus, a multimodal dataset with missing modalities can be formulated as $\tilde{D} = \{D^c, \tilde{D}^{m1}, \tilde{D}^{m2}\}$.

\begin{figure*}[t]
    \centering
    \includegraphics[width=1.03\linewidth]{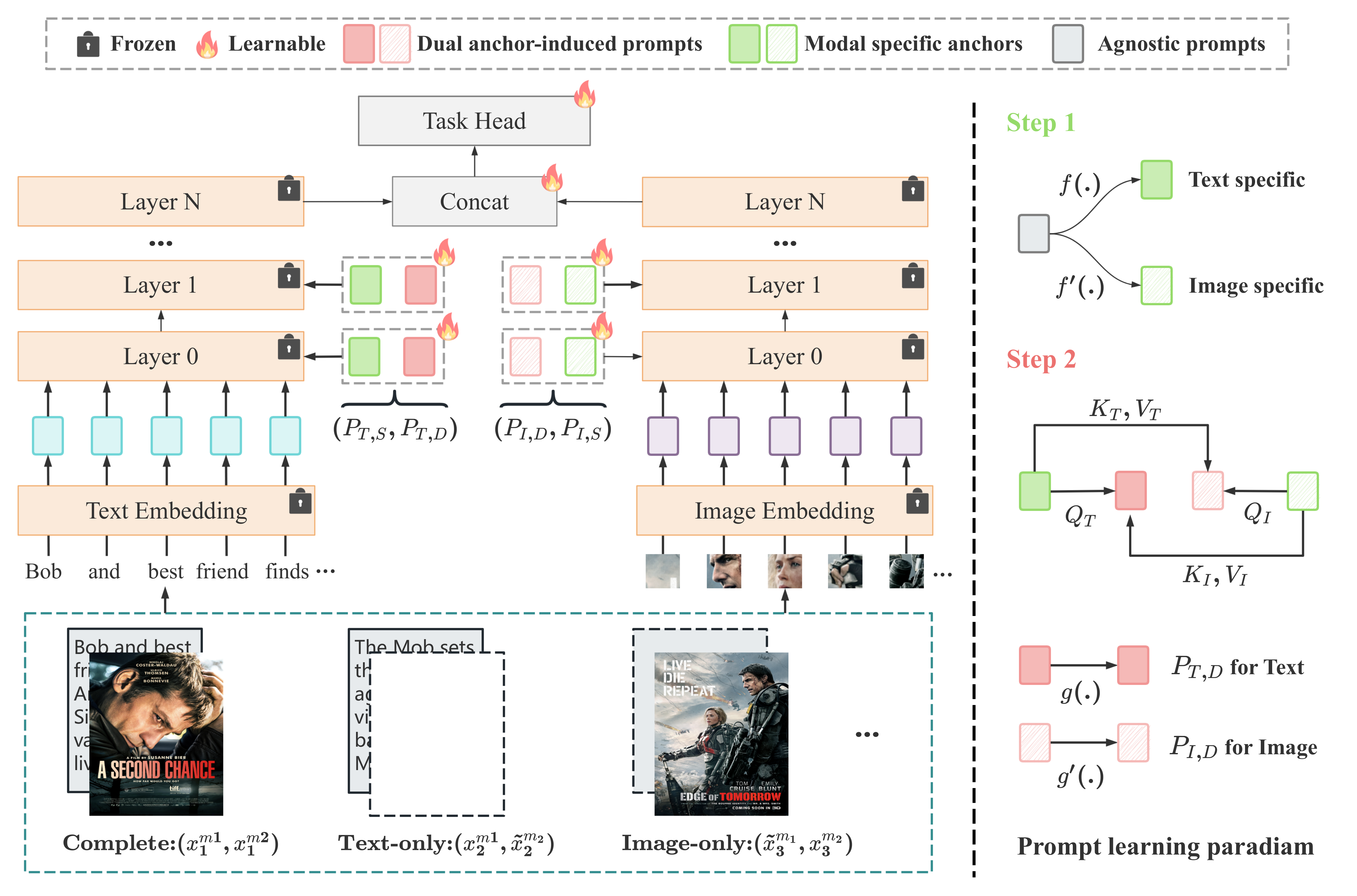}
    \caption{Overview of the proposed Learning from Reliable Latent Prompts (LLP) framework. Given multimodal inputs with potentially missing modalities, LLP introduces modality-specific latent anchors (i.e., learnable prompts) $P_{I, S}(P_{T, S})$ at each encoder layer. These anchors are input-agnostic and capture stable modality-level priors. A dual-path cross-attention mechanism enables anchors from different modalities to exchange complementary information and jointly generate dual anchor-induced prompts $P_{I, D}(P_{T, D})$. The refined prompts are integrated into both text and image encoders, and the resulting representations are fused for downstream prediction. This anchor-based prompting paradigm facilitates robust semantic compensation under missing-modality scenarios.}
    \label{fig:main}
    \vspace{-0.5cm}
\end{figure*}

Figure~\ref{fig:main} shows the overall framework of our proposed LLP. We adopt the widely-used two-stream multimodal architecture CLIP~\cite{clip} as our model backbone. We first encode the input text and image into token sequences using the pre-trained embedding layers of pretrained CLIP~\cite{clip}. For each input sample, we select the corresponding prompts \(P_m^I\) and \(P_m^T\) for the image and text encoders, respectively, according to the missing-modality type \(m \in \{c, m_1, m_2\}\). The prompt is composed of two types of prompts, including modality-specific latent prompts $P_{I, S} (P_{T, S})$, and dual anchor-induced prompts $P_{I, D} (P_{T, D})$. We prepend the prompts to the input tokens \(x_{m_1}\) and \(x_{m_2}\), forming unified input sequences for subsequent processing. We combine the task-related tokens from both encoders to form the final representation, which is then fed into a fully connected layer for prediction. During training, only the fully connected layer and the proposed prompts are optimized, while the backbone networks (i.e., the text and image embedding layers and encoders) remain frozen. We next present the detailed design of our prompting mechanism.

\subsection{Learning from Reliable Prompt learning}

\subsubsection{Modality-Specific Latent Anchors}

We first introduce modality-specific latent anchors, which are input-agnostic learnable prompt tokens that capture stable modality-level priors and serve as the foundation for subsequent prompt generation. Existing input-conditioned prompting methods~\cite{syp, dcp} rely on instance-level features that become unreliable under missing-modality settings, leading to unstable prompt learning and error accumulation. Instead of building upon such corrupted signals, we construct prompts from clean and informative latent anchors. These anchors are modeled as input-agnostic learnable prompts, which encode modality-level priors and have been shown to capture meaningful semantic information across different missing scenarios~\cite{maple, first_prompt}. This design ensures that prompt generation is grounded in stable and reliable representations.

Specifically, we derive modality-specific latent anchors from input-agnostic prompts for each encoder. The latent anchors for the image and text encoders are defined as:
\begin{equation}
    P_{I,S} = f_I(P_{I,A}), \quad P_{T,S} = f_T(P_{T,A}),
\end{equation}
where $P_{I,A}$ and $P_{T,A}$ denote the input-agnostic prompts for the image and text encoders, respectively. 

The transformation functions $f_I(\cdot)$ and $f_T(\cdot)$ share the same architecture, implemented as a fully connected layer (FC) with GELU activation~\cite{gelu} followed by Layer Normalization (LN)~\cite{layernorm}. Specifically, the function $f(\cdot)$ is formulated as:
\begin{equation}
    f(\cdot) = \mathrm{LN}\big(\mathrm{FC}(\mathrm{GELU}(\mathrm{FC}(\cdot)))\big).
\end{equation}

The latent anchor $P_{I,S}(P_{T,S})$ stores the stable latent information for each modality under different missing cases.

\subsubsection{Dual Anchor-induced Prompts}

Building upon the modality-specific latent anchors introduced above, we next describe how to generate refined prompts for robust cross-modal interaction. Instead of directly conditioning on unreliable input features, we derive prompts from the latent anchors through a dual-path cross-attention mechanism. Specifically, anchors from different modalities interact to exchange complementary information, enabling the model to capture cross-modal dependencies and synthesize modality-specific prompts. This design allows the generated prompts to remain stable and informative, providing reliable semantic compensation even under severe modality missing.

Building upon the modality-specific latent anchors, we generate refined prompts via a dual-path cross-attention mechanism. Specifically, given the latent anchors for text and image modalities, denoted as $P_{T,S}$ and $P_{I,S}$, we model cross-modal interactions by performing bidirectional cross-attention.

For the text branch, we use $P_{T,S}$ as queries and $P_{I,S}$ as keys and values:
\begin{equation}
    \tilde{P}_{T} = \mathrm{MHCA}(Q = P_{T,S},\; K = P_{I,S},\; V = P_{I,S}),
\end{equation}

Similarly, for the image branch, we use $P_{I,S}$ as queries and $P_{T,S}$ as keys and values:
\begin{equation}
    \tilde{P}_{I} = \mathrm{MHCA}(Q = P_{I,S},\; K = P_{T,S},\; V = P_{T,S}),
\end{equation}

where $\mathrm{MHCA}(\cdot)$ denotes multi-head cross-attention~\cite{attention}. Through this bidirectional interaction, the anchors exchange complementary information across modalities.

The final modality-specific prompts are obtained by applying a lightweight transformation with a bottleneck multi-layer perceptron (MLP) and Layer normalization~\cite{layernorm}:
\begin{equation}
    P_{I,D} = \mathrm{LN}(\mathrm{MLP}(\mathrm{LN}(\tilde{P}_{I}))),
\end{equation}
\begin{equation}
    P_{T,D} = \mathrm{LN}(\mathrm{MLP}(\mathrm{LN}(\tilde{P}_{T}))).
\end{equation}

By grounding prompt generation in input-agnostic latent anchors and modeling cross-modal interactions via dual-path attention, our approach mitigates noise introduced by incomplete inputs, facilitates effective information exchange, and yields robust and reliable semantic compensation even under extreme missing-modality scenarios. 

The final prompts $P_m^I$ and $P_m^T$ are constructed by concatenating the modality-specific latent anchors ($P_{I,S}, P_{T,S}$) and the dual anchor-induced prompts ($P_{I,D}, P_{T,D}$), respectively:
\begin{equation}
    P_m^I = [P_{I,S}, P_{I,D}], \quad  P_m^T = [P_{T,S}, P_{T,D}],
\end{equation}
where $[\cdot, \cdot]$ denotes the concatenation operation.

\subsubsection{Layer-wise Prompt Generation}

Following prior works~\cite{dcp, syp}, we adopt a layer-wise prompt generation strategy to maintain coherent information flow across encoder layers. Specifically, the generated prompts are injected into each layer and progressively refined along the network depth. For the image encoder, we denote the prompt at the $i$-th layer as $P_m^{I, R_i}$, where $i \in [1, N]$. The prompt is updated based on the previous layer via:
\begin{equation}
    P_m^{I, R_i} = \mathrm{LN}\big(\mathrm{FC}(\mathrm{GELU}(\mathrm{FC}(P_m^{I, R_{i-1}})))),
\end{equation}
where $\mathrm{FC}$ denotes a bottleneck MLP, $\mathrm{GELU}$~\cite{gelu} is the activation function, and $\mathrm{LN}$~\cite{layernorm} denotes Layer Normalization. This design enables the prompts to capture hierarchical representations across layers, enhancing their expressiveness for multimodal reasoning.

\section{Experiments}

\subsection{Experiment Settings}
\textbf{Datasets.}
We follow prior works~\cite{first_prompt, dcp, syp} and evaluate our method on three widely used benchmark datasets:

MM-IMDb~\cite{mmimdb} is a large-scale multimodal dataset for movie genre prediction, containing 25,959 movies annotated with both images and textual plot descriptions. Since each movie may belong to multiple genres, the task is formulated as a multi-label classification problem. This dataset evaluates the model’s ability to jointly reason over visual and textual modalities for semantic understanding.

UPMC Food-101~\cite{food-101} is a large multimedia dataset consisting of noisy image-text pairs collected from the web, covering 101 food categories. The textual descriptions are often unstructured and noisy, making the dataset particularly challenging for multimodal learning. It is widely used for food classification tasks requiring robust cross-modal alignment.

Hateful Memes~\cite{hatefulmemes} is a multimodal benchmark for hateful meme detection, containing over 10,000 image-text pairs. The dataset is carefully constructed such that neither modality alone is sufficient for accurate prediction, explicitly requiring joint multimodal reasoning. This makes it a challenging testbed for evaluating robustness under incomplete or misleading modality inputs.

\textbf{Implementation details.} We leverage CLIP~\cite{clip} as our multimodal backbone with a ViT-B/16~\cite{vit} image encoder. We resize input images to $224 \times 224$ and split them into patches of size $16 \times 16$. Text inputs are processed using the CLIP tokenizer with a maximum sequence length of 77. Both the image and text encoders are kept frozen during training, and only the prompt parameters, together with the task-specific fully connected (FC) layer are optimized. The feature dimensions for both modalities are set to 512. We use learnable prompts of length $L_p = 16$, which are injected into features from $M = 6$ layers. For missing modalities, we replace the corresponding inputs with zero-filled tensors to maintain a consistent input structure. We train the model using the AdamW optimizer with a learning rate of $1 \times 10^{-2}$ and a weight decay of $2 \times 10^{-2}$. The learning rate is linearly warmed up for 10\% of the total training steps and then decayed to zero. All experiments are conducted on a single NVIDIA RTX A6000 GPU with a batch size of 32.

\textbf{Missing Modality Setting.}  In this paper, we consider a realistic multimodal setting where modalities may be partially missing during both training and evaluation. Following prior works, we characterize the level of missing data by a missing rate $\eta$, defined as the fraction of samples with incomplete modalities. For vision-language tasks, we consider three missing scenarios: missing-text, missing-image, and missing-both. In the missing-text or missing-image setting, a proportion $\eta$ of the samples contain only a single modality, while the remaining $(1-\eta)$ samples retain both modalities. In the missing-both setting, the dataset is constructed such that $\frac{\eta}{2}$ of the samples are text-only, $\frac{\eta}{2}$ are image-only, and the remaining $(1-\eta)$ are modality-complete. This formulation can be extended to multimodal settings with more than two modalities, where each incomplete case is sampled proportionally based on $\frac{\eta}{M^2-2}$, while $(1-\eta)$ of the data remains fully observed.

\textbf{Baselines and Metrics.} We compare our proposed LLP with six competitive baselines: (1) CoOp~\cite{coop}, which learns input-level prompts; (2) MMP~\cite{first_prompt}, which introduces missing-aware prompts for both input and intermediate features; (3) MaPLe~\cite{maple}, which performs cross-modal prompt learning between image and text encoders; (4) DePT~\cite{dept}, which disentangles modality-specific knowledge into separate feature spaces; (5) DCP~\cite{dcp}, which models prompt correlations and inter-layer features interactions for enhanced representation learning; and (6) SyP~\cite{syp}, which employs an adapter to jointly learn from instance features and static prompts. Following prior works, we leverage F1-Macro to measure the multi-label classification performance on MM-IMDb dataset~\cite{mmimdb},  top-1 classification accuracy to evaluate the recognition performance on UPMC Food-101 dataset~\cite{food-101}, and Area Under the Receiver Operating Characteristic Curve (AUROC) for Hateful Memes~\cite{hatefulmemes}.

\subsection{Experiment results}

\begin{table}[t]
\centering
\caption{Comparison with CoOp~\cite{coop}, MMP~\cite{first_prompt}, MaPLe~\cite{maple}, DePT~\cite{dept}, DCP~\cite{dcp}, and SyP~\cite{syp} on the MM-IMDb~\cite{mmimdb}, UPMC Food-101~\cite{food-101}, and Hateful Memes~\cite{hatefulmemes} datasets under various missing-modality cases with different missing rates. The bold number indicates the best performance.}
\label{tab:results_testing}
\small 
\setlength{\aboverulesep}{0pt}
\setlength{\belowrulesep}{0pt}
\renewcommand{\arraystretch}{1.0} 

\begin{tabular}{l|c|cc|ccccccc} 
\toprule 
\multirow{2}{*}{Datasets} & \multirow{2}{*}{\begin{tabular}[c]{@{}c@{}}Missing\\ rate $\eta$\end{tabular}} & \multicolumn{2}{c|}{Train/Test}  & \multirow{2}{*}{CoOp} & \multirow{2}{*}{MMP} & \multirow{2}{*}{MaPLe} & \multirow{2}{*}{DePT} & \multirow{2}{*}{DCP} & \multirow{2}{*}{SyP} &  \multirow{2}{*}{\textbf{LLP}}\\  
` &  & Image & Text &  & &  &  &  & & \\ \hline

\multirow{9}{*}{\begin{tabular}[c]{@{}l@{}}MM-IMDb\\ (F1-Macro)\end{tabular}} 
 & \multirow{3}{*}{50\%} & 100\% & 50\% & 48.06 & 48.88 & 49.58 & 50.64 & 52.13 & 53.90 & \cellcolor{gray!15}\textbf{54.34}  \\
 &  & 50\% & 100\% & 49.89 & 51.46 & 52.32 & 52.78 & 54.32 & 56.27 & \cellcolor{gray!15}\textbf{56.87} \\
 &  & 75\% & 75\% & 48.37 & 49.32 & 49.56 & 50.87 & 52.32 & 55.02 & \cellcolor{gray!15}\textbf{55.89}\\ \cline{2-11} 
 & \multirow{3}{*}{70\%} & 100\% & 30\% & 44.13 & 45.64 & 45.52 & 46.38 & 48.52 & 51.37 & \cellcolor{gray!15}\textbf{52.20} \\
 &  & 30\% & 100\% & 48.82 & 50.52 & 50.64 & 52.13 & 53.14 & 54.20 & \cellcolor{gray!15}\textbf{55.65} \\
 &  & 65\% & 65\% & 46.84 & 48.12 & 49.16 & 50.32 & 51.42 & 52.90 &\cellcolor{gray!15}\textbf{53.28} \\ \cline{2-11} 
 & \multirow{3}{*}{90\%} & 100\% & 10\% & 44.76 & 45.32 & 46.84 & 47.56 & 49.26 & 50.21 &  \cellcolor{gray!15}\textbf{50.47}\\
 &  & 10\% & 100\% & 48.32 & 49.12 & 50.13 & 50.88 & 52.22 & 53.72 &\cellcolor{gray!15}\textbf{55.02} \\
 &  & 55\% & 55\% & 44.12 & 44.87 & 45.12 & 46.54 & 48.04 & 49.63 & \cellcolor{gray!15}\textbf{51.25} \\ \hline

\multirow{9}{*}{\begin{tabular}[c]{@{}l@{}}Food101\\ (Accuracy)\end{tabular}} 
 & \multirow{3}{*}{50\%} & 100\% & 50\% & 77.45 & 77.89 & 79.64 & 80.16 & 82.11 & 83.20 & \cellcolor{gray!15}\textbf{84.28} \\
 &  & 50\% & 100\% & 87.02 & 87.16 & 87.35 & 82.14 & 89.12 & 89.64 &\cellcolor{gray!15}\textbf{89.85} \\
 &  & 75\% & 75\% & 81.24 & 81.72 & 82.34 & 83.12 & 85.24 & 86.17 & \cellcolor{gray!15}\textbf{86.74} \\ \cline{2-11} 
 & \multirow{3}{*}{70\%} & 100\% & 30\% & 76.34 & 76.52 & 77.02 & 77.34 & 78.87 & 79.56 & \cellcolor{gray!15}\textbf{80.68} \\
 &  & 30\% & 100\% & 84.78 & 85.64 & 85.89 & 86.12 & 87.32 & 88.67 & \cellcolor{gray!15}\textbf{88.94} \\
 &  & 65\% & 65\% & 78.87 & 79.12 & 79.84 & 81.46 & 81.87 & 82.45 &\cellcolor{gray!15}\textbf{83.22} \\ \cline{2-11} 
 & \multirow{3}{*}{90\%} & 100\% & 10\% & 71.87 & 73.14 & 73.46 & 74.12 & 75.26 & 76.33 & \cellcolor{gray!15}\textbf{78.10} \\
 &  & 10\% & 100\% & 81.67 & 82.14 & 83.12 & 83.56 & 85.78 & 86.41 & \cellcolor{gray!15}\textbf{86.83} \\
 &  & 55\% & 55\% & 76.46 & 76.58 & 77.85 & 78.12 & 79.87 & 81.03 & \cellcolor{gray!15}\textbf{81.69} \\ 
 \midrule

\multirow{9}{*}{\begin{tabular}[c]{@{}l@{}}Hateful\\ Memes\\ (AUROC)\end{tabular}} 
 & \multirow{3}{*}{50\%} & 100\% & 50\% & 60.56 & 60.31 & 60.87 & 61.87 & 62.32 & 68.25 & \cellcolor{gray!15}\textbf{70.84} \\
 &  & 50\% & 100\% & 62.41 & 62.35 & 63.13 & 63.88 & 64.46 & 66.80 & \cellcolor{gray!15}\textbf{68.07} \\
 &  & 75\% & 75\% & 64.87 & 65.84 & 65.46 & 65.86 & 66.02 & 68.16 & \cellcolor{gray!15}\textbf{69.38} \\ \cline{2-11} 
 & \multirow{3}{*}{70\%} & 100\% & 30\% & 60.74 & 61.12 & 61.26 & 61.56 & 62.82 & 68.94 & \cellcolor{gray!15}\textbf{70.28} \\
 &  & 30\% & 100\% & 62.74 & 63.24 & 63.14 & 63.48 & 64.12 & 66.98 & \cellcolor{gray!15}\textbf{67.25} \\
 &  & 65\% & 65\% & 64.82 & 65.04 & 65.23 & 65.48 & 66.08 & 68.42 & \cellcolor{gray!15}\textbf{69.62} \\ \cline{2-11} 
 & \multirow{3}{*}{90\%} & 100\% & 10\% & 60.03 & 57.21 & 60.74 & 61.14 & 62.08 & 69.70 & \cellcolor{gray!15}\textbf{70.90} \\
 &  & 10\% & 100\% & 61.46 & 61.52 & 61.87 & 62.42 & 63.87 & 64.54 &\cellcolor{gray!15}\textbf{65.19} \\
 &  & 55\% & 55\% & 64.32 & 63.34 & 64.85 & 65.37 & 66.78 & 68.93 & \cellcolor{gray!15}\textbf{69.17} \\ 
 \bottomrule
\end{tabular}
\end{table}

\textbf{Comparison with the State-of-the-arts.} We first conduct experiments on three widely-used datasets: MM-IMDb~\cite{mmimdb}, UPMC
Food-101~\cite{food-101}, and Hateful Memes~\cite{hatefulmemes}. we compare our proposed LLP method with six state-of-the-art methods, including: (1) CoOp~\cite{coop}, (2) MMP~\cite{first_prompt},  (3) MaPLe~\cite{maple}, (4) DePT~\cite{dept}, (5) DCP~\cite{dcp}, and (6) SyP~\cite{syp}. We evaluate these methods across three different missing rates, including $\eta=50\%$, $\eta=70\%$, and $\eta=90\%$, upon three various missing-modality cases, including missing-image, missing-text and missing-both, in Table~\ref{tab:results_testing}.

As shown in Table~\ref{tab:results_testing}, LLP consistently outperforms state-of-the-art methods, including SyP~\cite{syp} and DCP~\cite{dcp}, across all missing-modality settings. The performance gains are particularly pronounced under high missing rates ($\eta = 90\%$), demonstrating the effectiveness of the latent prior-driven prompting paradigm in improving the robustness of multimodal models. The gains are particularly pronounced on more challenging benchmarks. On Hateful Memes~\cite{hatefulmemes}, which requires strong cross-modal reasoning, LLP achieves significant improvements, highlighting its ability to capture complementary multimodal information. Similarly, on MM-IMDb~\cite{mmimdb}—a difficult multi-label classification task with relatively low overall performance—LLP consistently outperforms all baselines across different missing settings. Despite the performance being in the 40–50 F1 range, LLP still delivers stable and noticeable gains, demonstrating its effectiveness in handling complex and noisy multimodal scenarios.  These results suggest that the advantage of LLP is not dataset-specific, but generalizes well to challenging multimodal learning problems.

An interesting observation is that different datasets exhibit asymmetric sensitivity to modality missing. Specifically, MM-IMDb~\cite{mmimdb} and Food101~\cite{food-101} suffer more performance degradation when the text modality is missing, while Hateful Memes~\cite{hatefulmemes} is more sensitive to missing visual information. Despite these varying modality dependencies, LLP consistently achieves larger performance gains in the more challenging missing cases for each dataset (i.e., missing text for MM-IMDb~\cite{mmimdb} and Food101~\cite{food-101}, and missing images for Hateful Memes~\cite{hatefulmemes}).  This suggests that, unlike prior input-conditioned methods that rely on potentially incomplete or corrupted inputs, LLP effectively leverages modality-specific latent anchors as stable sources of information. By grounding prompt generation in these input-independent representations, LLP enables more reliable cross-modal compensation, leading to improved robustness across diverse missing-modality scenarios.

     

\begin{figure}[t]
    \centering
    \includegraphics[width=0.35\linewidth]{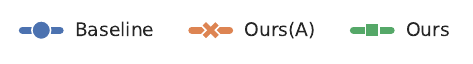}
    \vspace{-4mm}

    \subfigure[Missing Text]{
        \includegraphics[width=0.31\columnwidth]{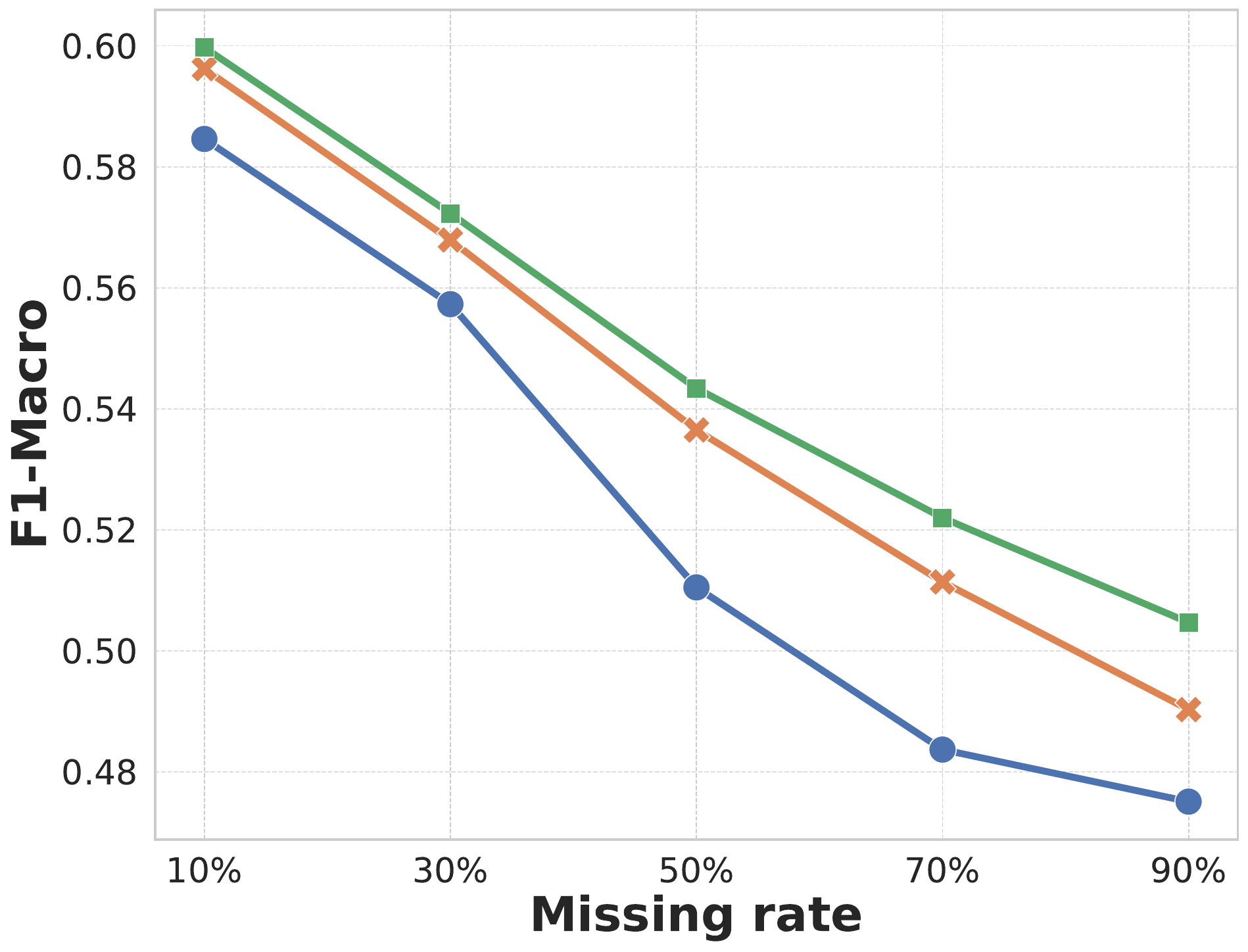}
    }
    \hfill 
    \subfigure[Missing Image]{
        \includegraphics[width=0.31\columnwidth]{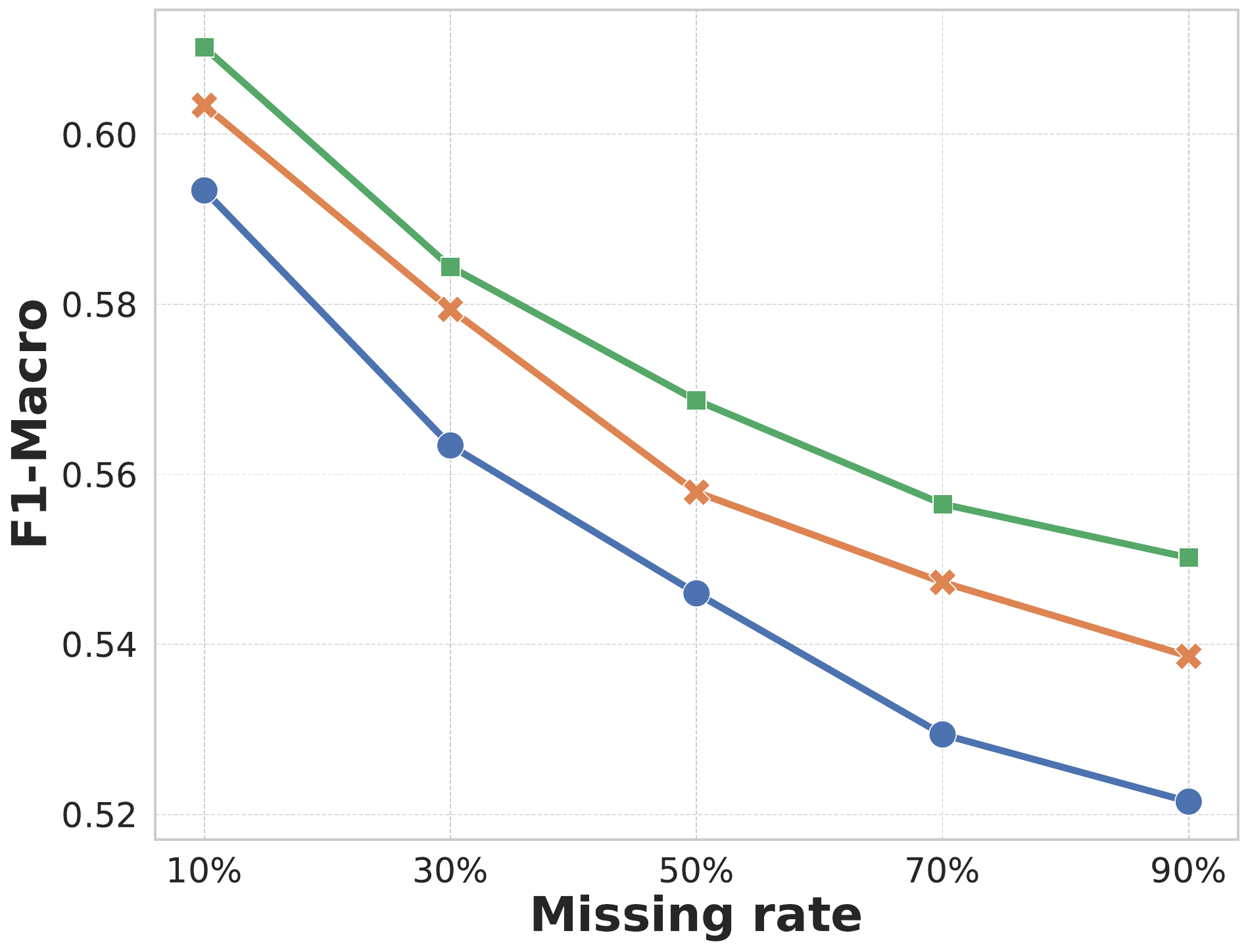}
    }
    \hfill
    \subfigure[Missing Both]{
        \includegraphics[width=0.31\columnwidth]{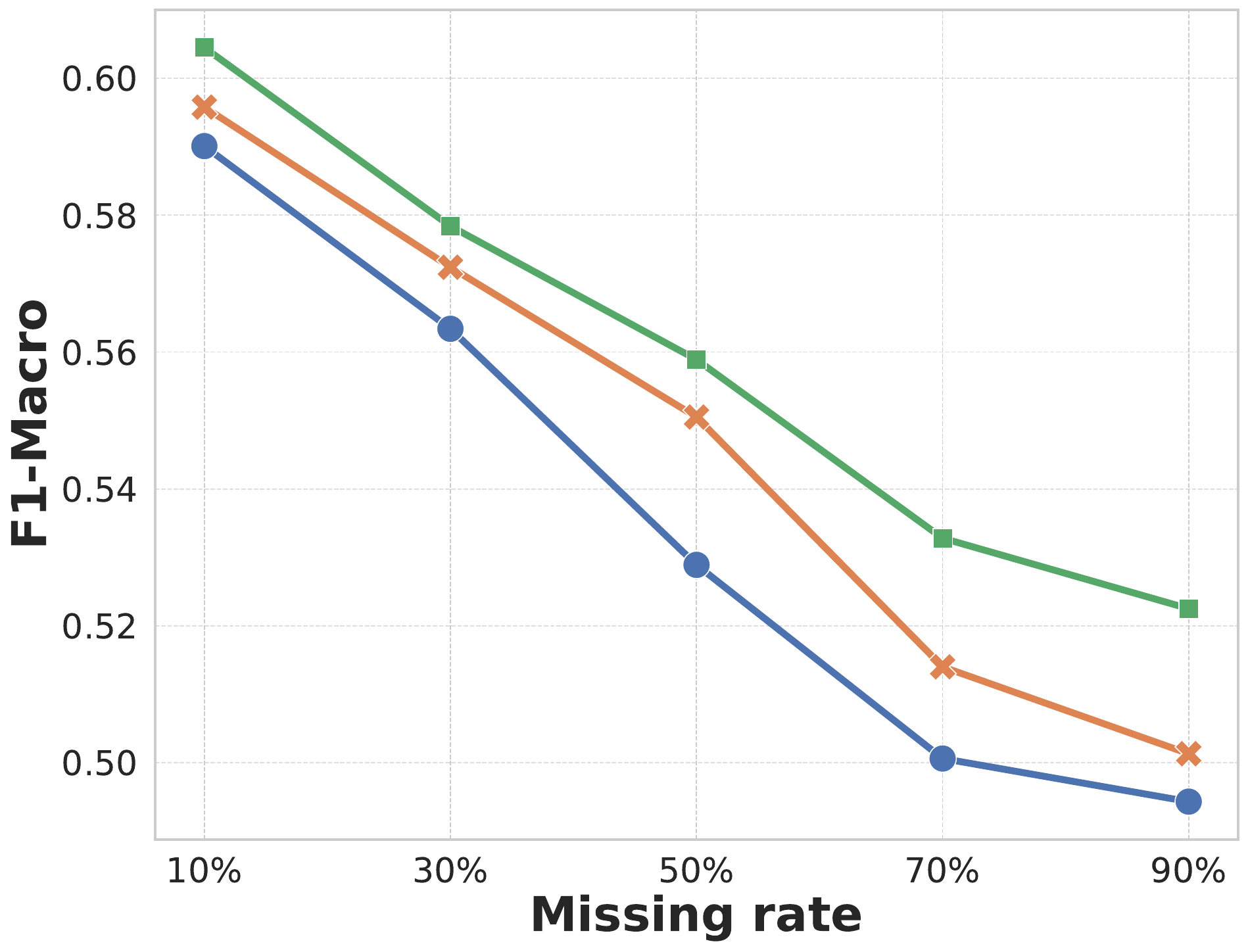}
    }
    
    \caption{Comparison of our final model (Ours) with (1) baseline, which directly drops the features when a modality is missing; (2) Ours (A), which only equips the Modality-Specific Latent Anchors. The experiments are conducted on the MM-IMDb dataset~\cite{mmimdb} across different missing rates under three missing-modality scenarios, with all models trained and evaluated at the same missing rate.} 
    \label{fig:ablation}
\end{figure}

\begin{figure}[t]
\centering
\subfigure[Missing Text]{
    \includegraphics[width=0.31\columnwidth]{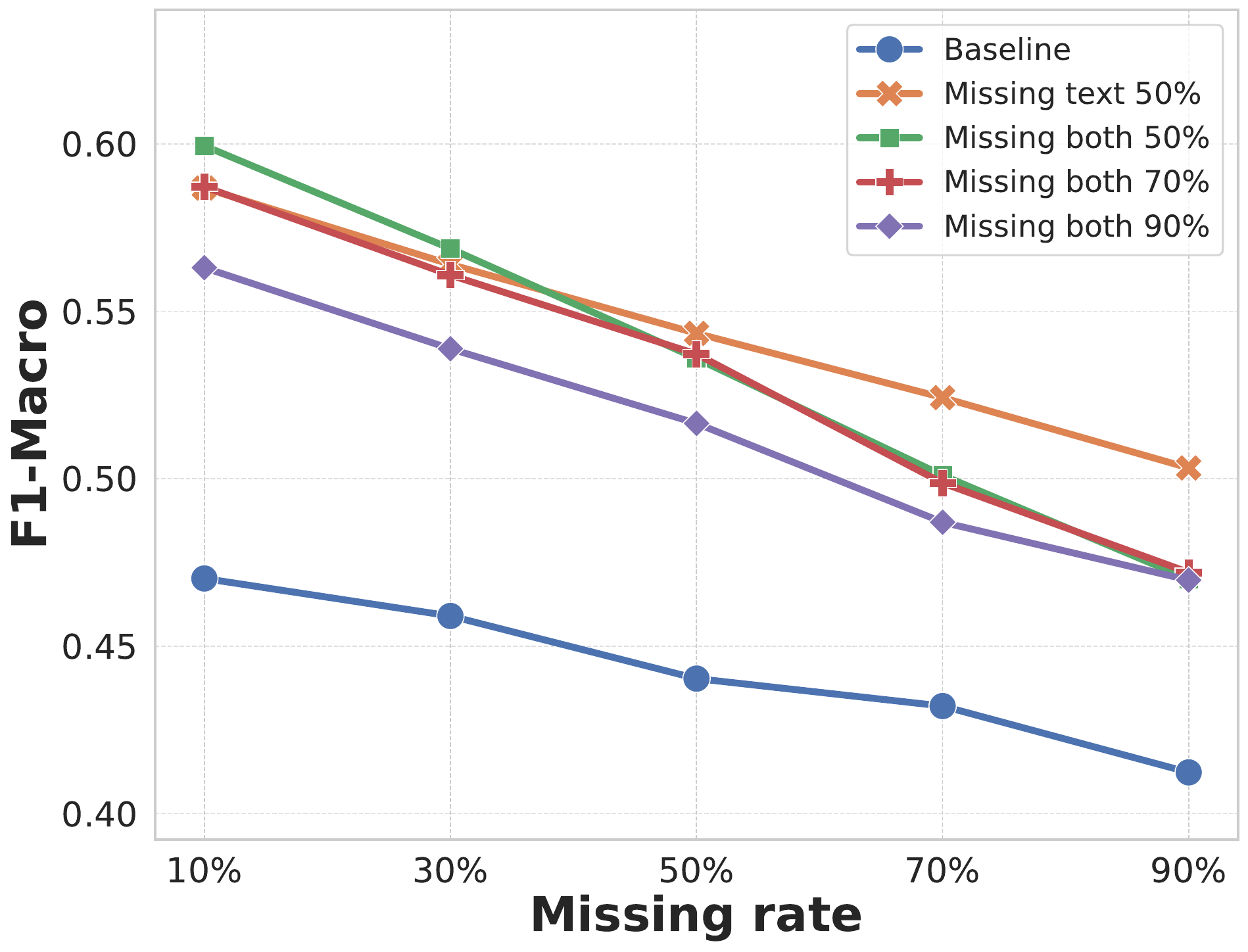}
}
\hfill
\subfigure[Missing Image]{
    \includegraphics[width=0.31\columnwidth]{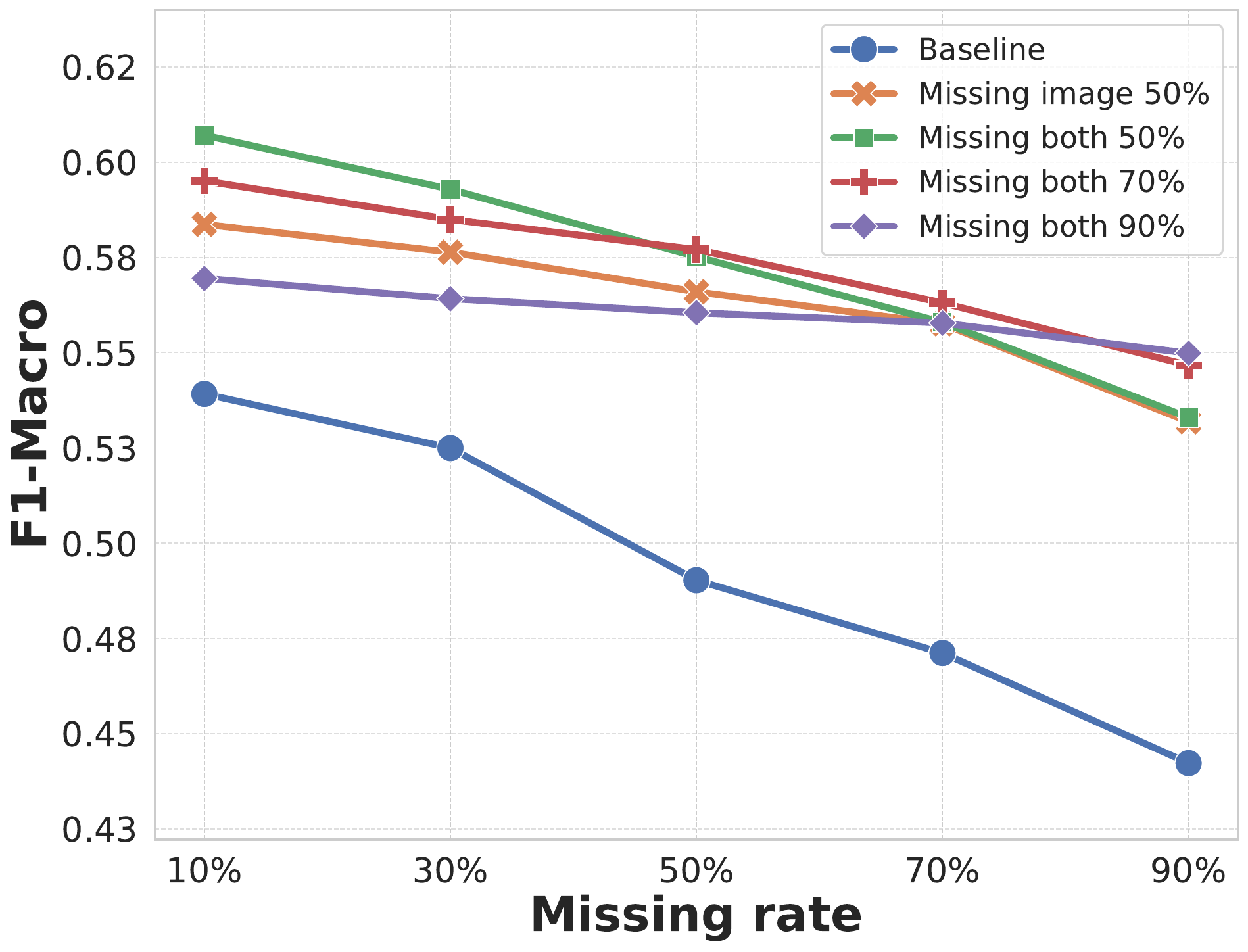}
}
\hfill
\subfigure[Missing Both]{

    \includegraphics[width=0.31\columnwidth]{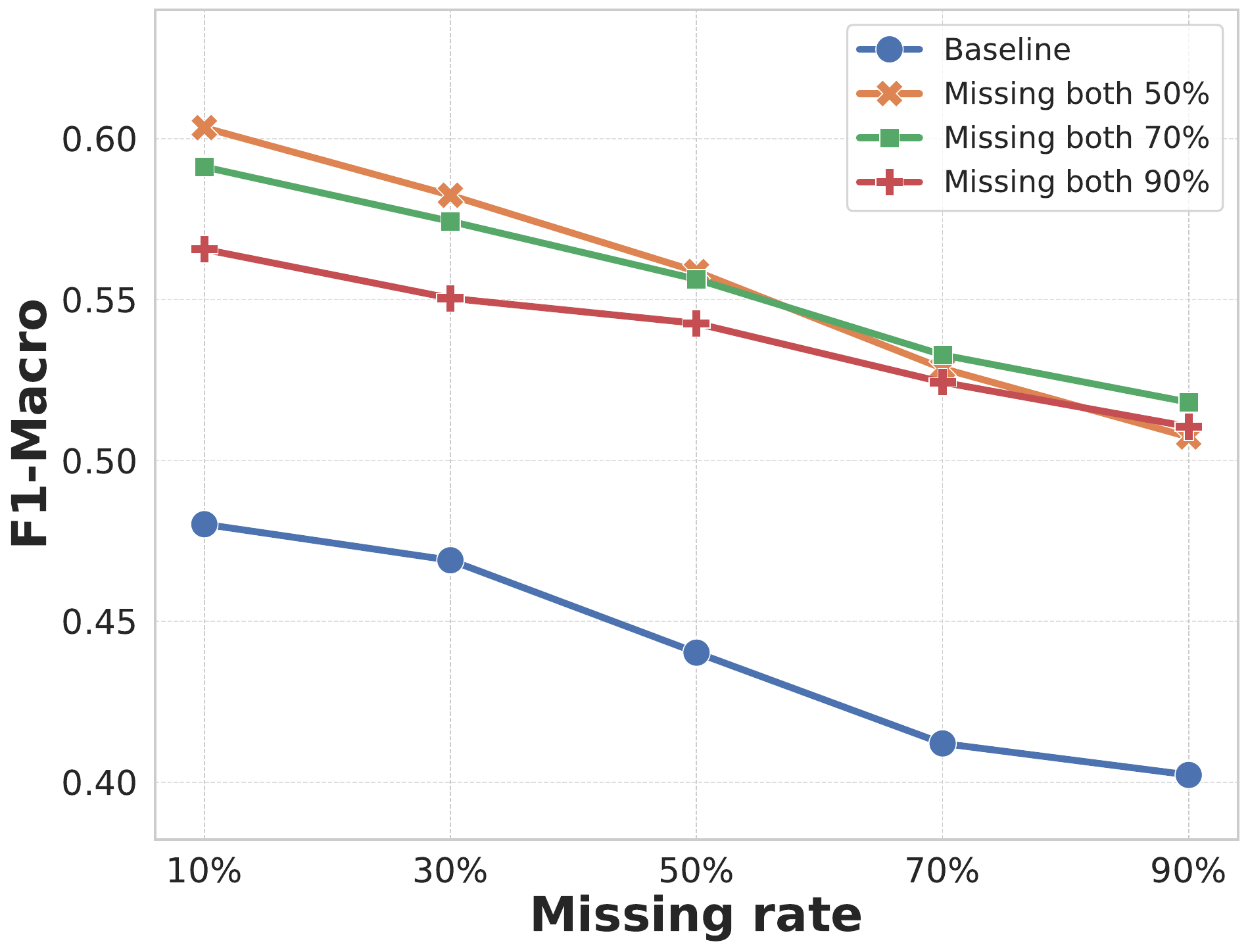}
    
}
    \caption{Generalizability Analysis of Our Method to Different Missing Rates on MM-IMDb dataset~\cite{mmimdb}. (a) All Models are trained on missing-both or missing-text cases, and evaluated on missing-text cases with different missing rates. (b) All models are trained on missing-both or missing-image cases, and evaluated on missing-image cases with different missing rates. (c) All models are trained on missing-both cases, and evaluated on missing-both cases with different missing rates.}
    \label{fig:generlization}
\end{figure}

\textbf{Ablation Studies.} We also conduct numerous ablation studies to evaluate the effectiveness of the proposed components of LLP. As shown in Figure~\ref{fig:ablation}, we first verify the effectiveness of our proposed components across different missing cases on the MM-IMDb dataset~\cite{mmimdb}, including missing-image, missing-text, and missing-both. Each model variant was trained and evaluated with a consistent missing rate. We first observe that all our variants consistently outperform the baseline across different missing-modality settings, demonstrating improved robustness under incomplete multimodal inputs. Notably, introducing modality-specific latent anchors alone already leads to substantial performance gains over the baseline, indicating that the learned anchors provide more stable and reliable modality-level representations. Building upon this, further incorporating dual anchor-induced prompts yields additional significant improvements, achieving the best overall performance. The results of Table~\ref{tab:more} further validate this point. This confirms the effectiveness of our design, where latent anchors serve as robust information sources that facilitate more reliable cross-modal interaction and semantic compensation. Moreover, the performance gap between the baseline, the latent-anchor-only variant, and the full model becomes increasingly pronounced as the missing rate grows. This trend highlights the superior robustness of our approach, especially under extreme missing-modality scenarios.

\begin{wraptable}{t}{0.53\textwidth}
\vspace{-8pt}
\begin{minipage}{0.49\textwidth}
\centering
\small
\caption{Additional ablation results for each component across datasets under missing-both ($\eta=70\%$). Ours (A), which only equips the Modality-Specific Latent Anchors.} 
\begin{tabular}{c | c c c}
\toprule
Variants  & MM-IMDb & Food101 & Hateful Memes \\
\midrule 
Ours (A) & 51.40\% & 81.76\% & 66.81\% \\
Ours     & \textbf{53.28\%} & \textbf{83.22\%} & \textbf{69.62\%} \\
\bottomrule
\end{tabular}
\label{tab:more}
\end{minipage}
\vspace{-8pt}
\end{wraptable}

\textbf{Generalization Capbility.} To assess the generalization ability of LLP under varying missing-modality conditions, we train models with different missing rates and evaluate them across three scenarios: missing text, missing image, and missing both modalities on the MM-IMDb dataset. The results are presented in Figure~\ref{fig:generlization}. We observe that all variants of LLP consistently outperform the baselines across a wide range of missing rates (10\%–90\%), demonstrating strong robustness to incomplete multimodal inputs. Moreover, models trained under single-modality missing settings exhibit competitive performance when evaluated on the corresponding missing scenario, indicating effective adaptation to modality-specific absence. Notably, models trained under the missing-both setting achieve strong and stable performance across all evaluation scenarios. This suggests that jointly modeling multiple missing patterns enables more flexible and robust behavior. Overall, LLP demonstrates superior generalization when trained under the missing-both condition, effectively handling diverse missing-modality scenarios in a stable and reliable manner.


\textbf{Robustness.} We further evaluate the robustness of LLP under varying missing rates (see \textbf{Appendix~\ref{ab:sec3}} for details). We observe that all baseline methods suffer from significant performance degradation as the missing rate increases, highlighting their sensitivity to incomplete multimodal inputs. In contrast, LLP consistently exhibits the smallest performance drop across different missing scenarios, and even shows performance improvements under high missing rates. These results demonstrate the effectiveness of our latent anchor-based design, which provides stable modality-level priors and enables reliable cross-modal compensation through dual anchor-induced prompts.

\textbf{Visualization Analysis.} We visualize prompt-enhanced features under the missing-both setting ($\eta = 90\%$). As shown in \textbf{Appendix~\ref{sec:ab1}}, SyP~\cite{syp} yields fragmented representations, where same-class samples are split into modality-specific clusters and lack consistent class-level structure. In contrast, LLP produces more compact intra-class distributions and clearer inter-class separation, indicating improved cross-modal alignment via latent anchors.

\textbf{Limitations and Discussion.} Despite the promising results, our work has several limitations. First, we focus on bimodal settings (image and text); extending LLP to more modalities (e.g., audio or video) remains future work. Second, our method is evaluated on commonly-used two-stream architectures~\cite{dcp, syp, maple}, and its applicability to other multimodal models (e.g., unified or large-scale foundation models) is not explored. Beyond these limitations, LLP offers several broader impacts. It provides a new prompting paradigm for handling missing modalities, a common challenge in real-world applications. In addition, LLP achieves strong performance with high parameter efficiency by optimizing only a small set of learnable parameters, making it practical for deployment. Our findings also demonstrate the promise of learning input-independent latent prompts.


\section{Concluding Remarks}
In this paper, we address the missing-modality problem in multimodal learning by proposing Learning from Reliable Latent Prompts (LLP), a latent prior-driven prompting paradigm that decouples prompt learning from unreliable input observations. By introducing modality-specific latent anchors and dual anchor-induced prompts, LLP enables stable prompt generation and effective cross-modal interaction, providing reliable semantic compensation even under severe modality missing. Extensive experiments across multiple benchmarks demonstrate that LLP consistently outperforms state-of-the-art methods under diverse missing-modality settings. Notably, the performance gains become more pronounced as the missing rate increases, and LLP shows strong robustness across both modality-specific and balanced missing scenarios. Further analyses, including ablation and generalization studies, validate the effectiveness of each component and highlight the importance of stable latent priors for robust multimodal reasoning. Overall, LLP offers a simple yet effective solution for handling missing modalities, providing a promising direction for robust and generalizable multimodal learning.



\section{References}

{
\small
    \bibliographystyle{abbrvnat}
    \bibliography{main}
}






\newpage
\appendix

\section{Technical appendices and supplementary material}
\label{appendix}

\begin{figure}[h]
    \centering
    \subfigure[SyP]{
        \includegraphics[width=50mm, height=40mm]{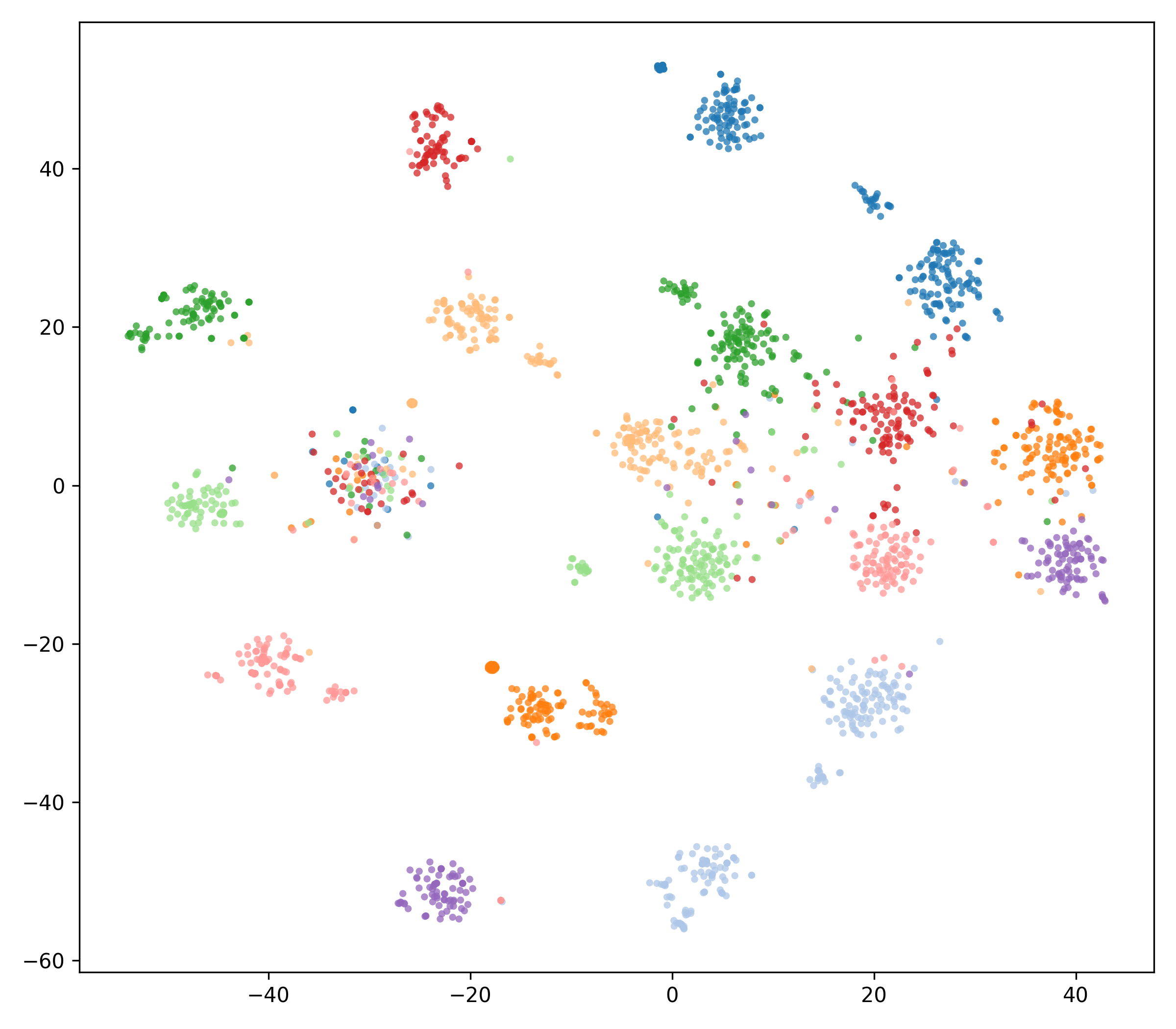}
    }
    \hspace{1cm}
    \subfigure[Ours]{
        \includegraphics[width=50mm, height=40mm]{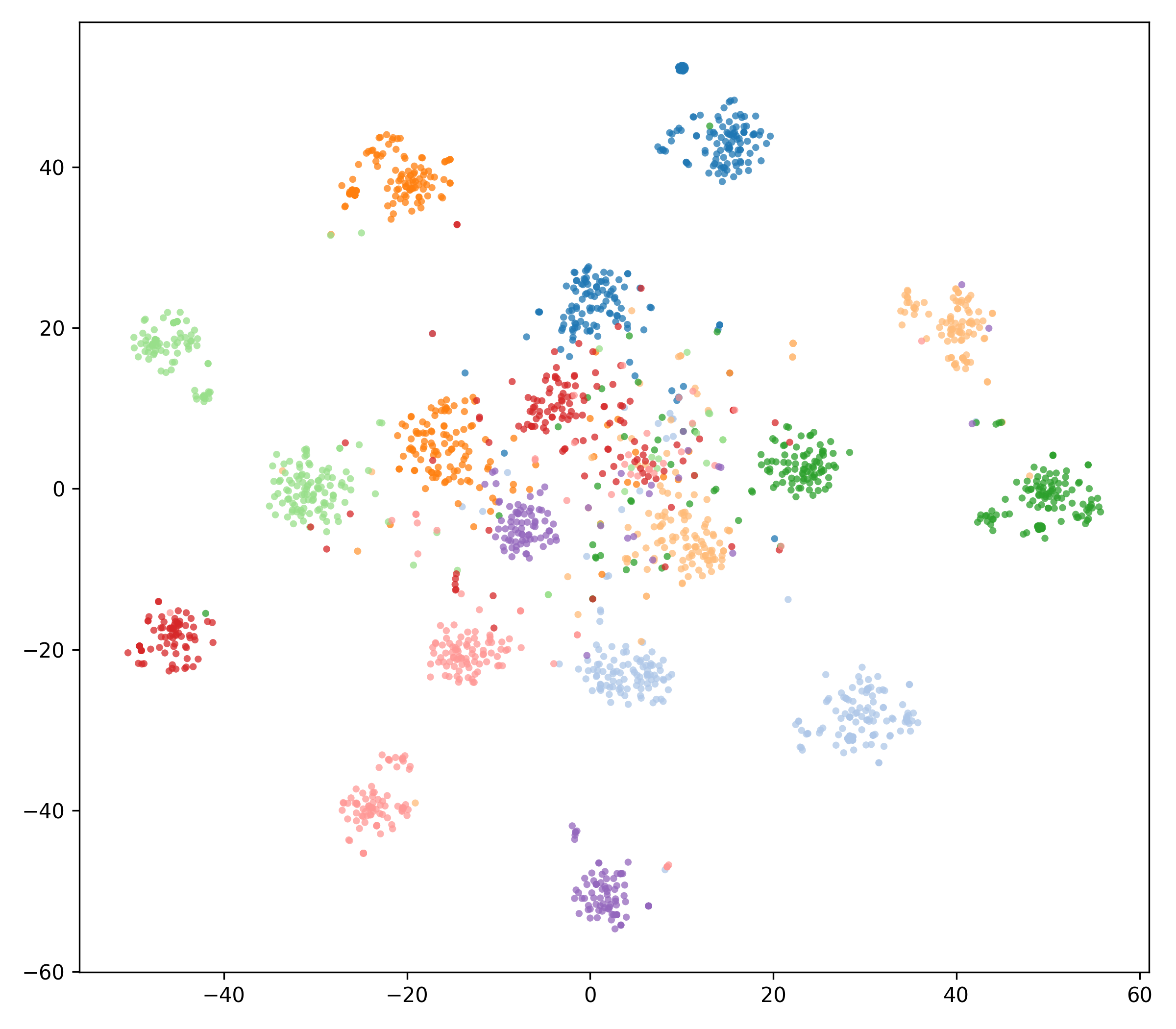}
    }
    
    \caption{Visualization of concatenated prompt-enhanced features from the two encoders on the Food101 dataset~\cite{food-101} under the missing-both setting with a high missing rate ($\eta = 90\%$). Compared to SyP~\cite{syp}, our method produces more compact intra-class distributions and clearer inter-class separation, indicating more consistent and robust representations under severe modality missing.} 
    \label{fig:proofs}
\end{figure}

\subsection{Visualization Study} 
\label{sec:ab1}

To better understand the effectiveness of our method, we visualize the concatenated prompt-enhanced features from the two encoders with a high missing rate ($\eta = 90\%$), as shown in Figure~\ref{fig:proofs}. We observe that SyP~\cite{syp}, which relies on input-conditioned dynamic prompts, tends to produce fragmented representations for each class. Specifically, samples from the same class are split into two distinct clusters. This phenomenon arises because, under the missing-both setting, representations are constructed from different incomplete sources, namely prompts combined with image features or prompts combined with text features. Due to the unreliability of input-conditioned features at high missing rates, these clusters are often significantly separated and may even overlap with clusters from other classes. As a result, while individual clusters remain locally separable, the overall class-level separability is degraded.

In contrast, our method produces more coherent and compact class distributions. The two clusters corresponding to the same class are much closer in the feature space, and the boundaries between different classes are better preserved. This indicates that LLP effectively aligns representations across different missing-modality conditions. 

We attribute this improvement to the use of modality-specific latent anchors, which provide stable and input-independent representations for prompt generation. By avoiding direct reliance on incomplete input features, LLP enables more reliable cross-modal compensation and maintains consistent semantic structures even under extreme missing conditions.

\subsection{Theoretical Analysis} 
\label{sec2}

We provide a theoretical perspective to understand the limitation of input-conditioned prompting under missing-modality settings. Consider a multimodal input $x = (x_I, x_T)$ and a missing mask $m \in \{0,1\}^2$, where $m_I = 0$ (or $m_T = 0$) indicates the absence of the image (or text) modality. Input-conditioned prompts are generated as $P = f(x \odot m)$, where $\odot$ denotes element-wise masking.

Let $\mathcal{L}$ denote the training objective. The gradient with respect to the prompt parameters $\theta$ can be written as:
\begin{equation}
\nabla_\theta \mathcal{L}
= \mathbb{E}_m \left[
\frac{\partial \mathcal{L}}{\partial P}
\cdot
\frac{\partial P}{\partial \theta}
\right],
\quad \text{where } P = f(x \odot m; \theta).
\end{equation}

Under standard missing-modality sampling, where the missing rate is $\eta$, the expected gradient can be decomposed as:
\begin{equation}
\mathbb{E}[\nabla_{\theta} \mathcal{L}] 
= (1 - \eta)\,\nabla_{\text{multi}} 
+ \eta\,\nabla_{\text{uni}},
\end{equation}
where $\nabla_{\text{multi}}$ and $\nabla_{\text{uni}}$ denote gradients from multimodal and unimodal observations, respectively.

As $\eta$ increases, the contribution from multimodal observations diminishes, and the gradient becomes increasingly dominated by unimodal signals:
\begin{equation}
\lim_{\eta \to 1} \mathbb{E}[\nabla_{\theta} \mathcal{L}] \approx \nabla_{\text{uni}}.
\end{equation}

This indicates that input-conditioned prompts are optimized primarily based on unimodal observations under high missing rates, resulting in a biased estimation of multimodal semantics. Consequently, the learned representations become modality-dependent and fail to preserve consistent cross-modal alignment, consistent with our empirical observations in Figure~\ref{fig:proofs}.

In contrast, LLP generates prompts from modality-specific latent anchors $P_S$, which are input-independent:
\begin{equation}
P = g(P_S; \theta),
\end{equation}
where $P_S$ does not depend on the input $x$ or the missing mask $m$. The corresponding gradient is:
\begin{equation}
\nabla_\theta \mathcal{L}
= \mathbb{E}_m \left[
\frac{\partial \mathcal{L}}{\partial P}
\cdot
\frac{\partial P}{\partial \theta}
\right],
\end{equation}
which is independent of the corrupted input $x \odot m$. Unlike input-conditioned prompting, LLP removes the dependency of prompt learning on incomplete observations. Therefore, its optimization is not biased toward unimodal signals even under high missing rates. This leads to more stable training dynamics and preserves consistent cross-modal alignment, providing a theoretical explanation for the superior robustness of LLP observed in our experiments.

\begin{figure}[h]
    \centering
    \includegraphics[width=0.75\linewidth]{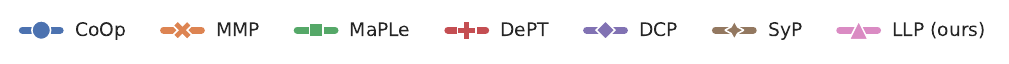}
    \vspace{-4mm}

    \subfigure[Missing Text]{
        \includegraphics[width=0.31\columnwidth]{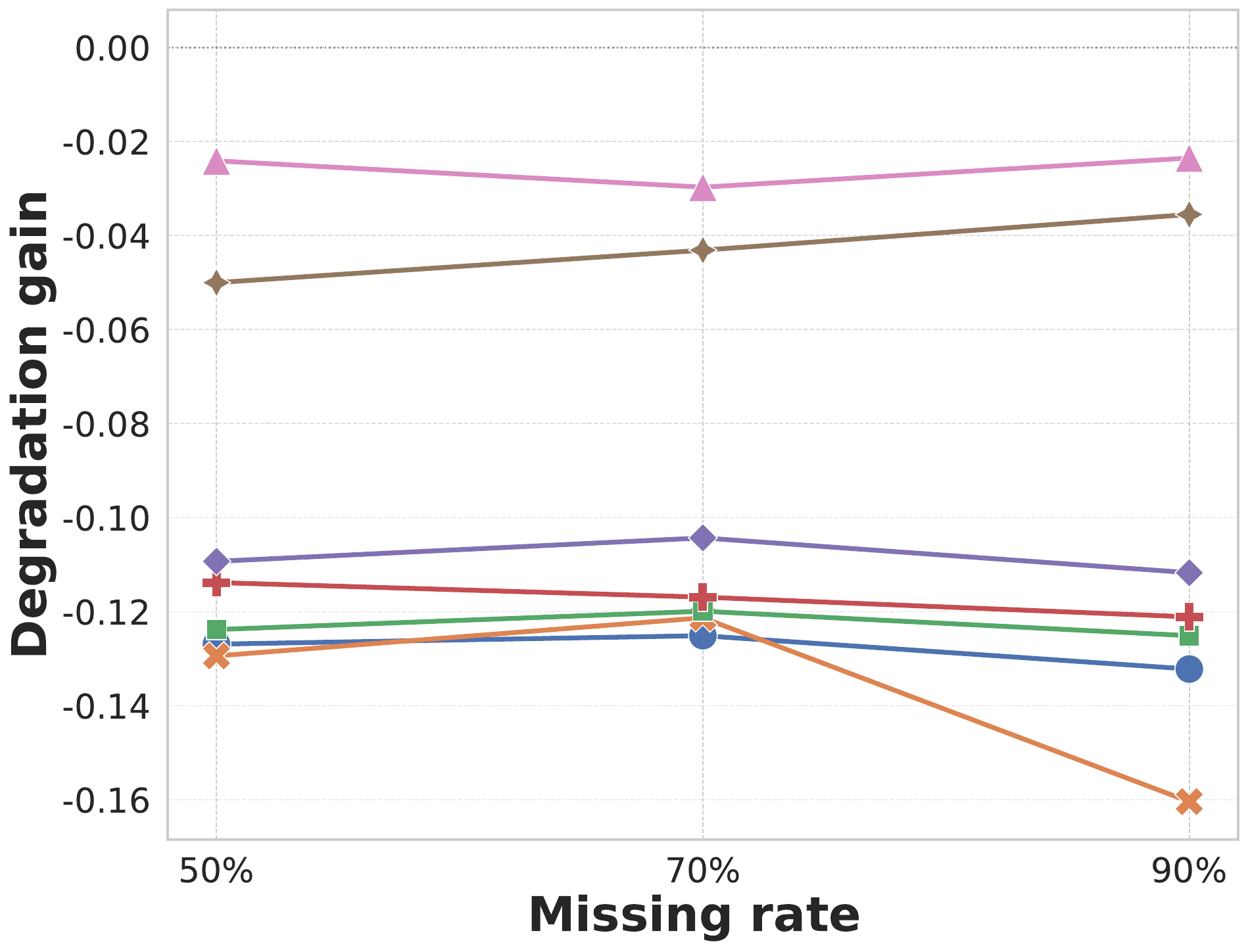}
    }
    \hfill 
    \subfigure[Missing Image]{
        \includegraphics[width=0.31\columnwidth]{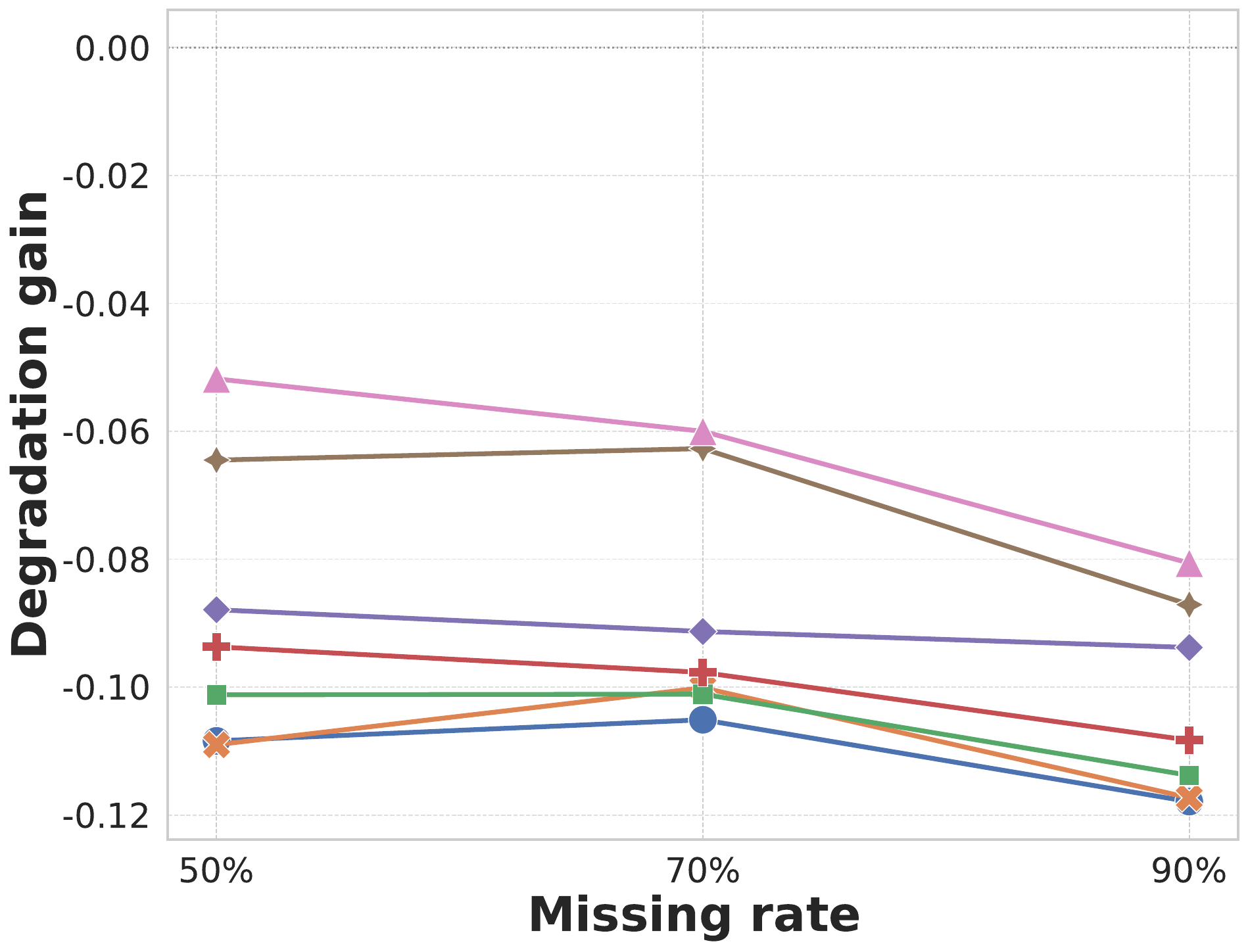}
    }
    \hfill
    \subfigure[Missing Both]{
        \includegraphics[width=0.31\columnwidth]{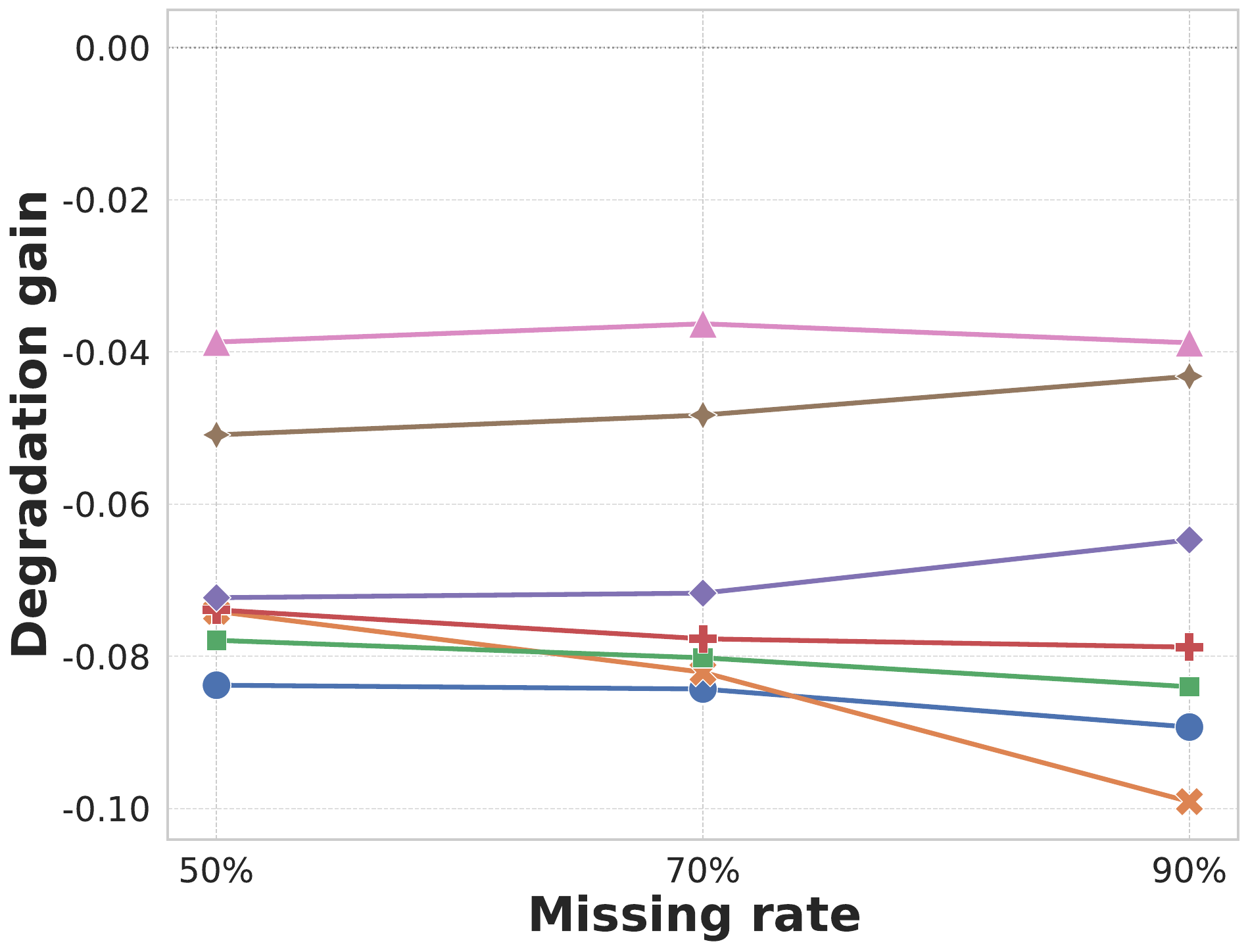}
    }
    
    \caption{Analysis of Robustness to Different Missing Rates on Hateful Memes dataset~\cite{hatefulmemes}. Notably, LLP maintains stable performance and even exhibits slight improvements under extreme missing conditions ($\eta = 90\%$), highlighting its effectiveness in mitigating missing-modality effects through the proposed latent anchor-based design.} 
    \label{fig:overall_robustness}
    \vspace{-0.5cm}
\end{figure}

\subsection{Robustness Analysis} 
\label{ab:sec3}

We provide a detailed analysis of model robustness under varying missing rates on the Hateful Memes dataset~\cite{hatefulmemes}, as illustrated in Figure~\ref{fig:overall_robustness}. Specifically, we evaluate different methods under three missing-modality scenarios: missing text, missing image, and missing both modalities, with missing rates ranging from 50\% to 90\%. Overall, all baseline methods exhibit a clear trend of performance degradation as the missing rate increases across all scenarios. This behavior highlights their strong dependence on input-conditioned features, which become increasingly unreliable when a larger proportion of modalities is absent. In particular, methods such as MMP and MaPLe show sharper performance drops at higher missing rates, indicating limited robustness under severe modality incompleteness.

In contrast, LLP consistently demonstrates superior robustness across all settings. As shown in Figure~\ref{fig:overall_robustness}, LLP achieves the smallest performance degradation in all three scenarios. Notably, under extreme missing conditions (e.g., $\eta=90\%$), LLP not only maintains stable performance but in some cases even shows slight improvements. This phenomenon suggests that LLP is less reliant on potentially corrupted input features and can instead leverage more stable internal representations. Furthermore, LLP exhibits consistent behavior across different types of missing modalities. Whether the missing information lies in the text, image, or both modalities, LLP maintains a relatively flat degradation curve compared to other methods. This indicates that LLP is capable of adapting to diverse missing-modality patterns without overfitting to any specific condition. The observed robustness can be attributed to the proposed latent anchor-based design. Unlike input-conditioned prompting methods, LLP introduces modality-specific latent anchors that serve as input-independent and stable representations of modality-level priors. These anchors provide a reliable foundation for prompt generation, especially when input features are incomplete or noisy. In addition, the dual anchor-induced prompting mechanism enables effective cross-modal interaction, allowing the model to compensate for missing information by leveraging complementary signals from other modalities. Overall, these results demonstrate that LLP significantly mitigates the impact of missing modalities and achieves strong robustness under varying missing rates. This further validates the effectiveness of learning from reliable latent prompts for handling incomplete multimodal data.

\begin{figure}[h]
    \centering
    \includegraphics[width=0.6\columnwidth]{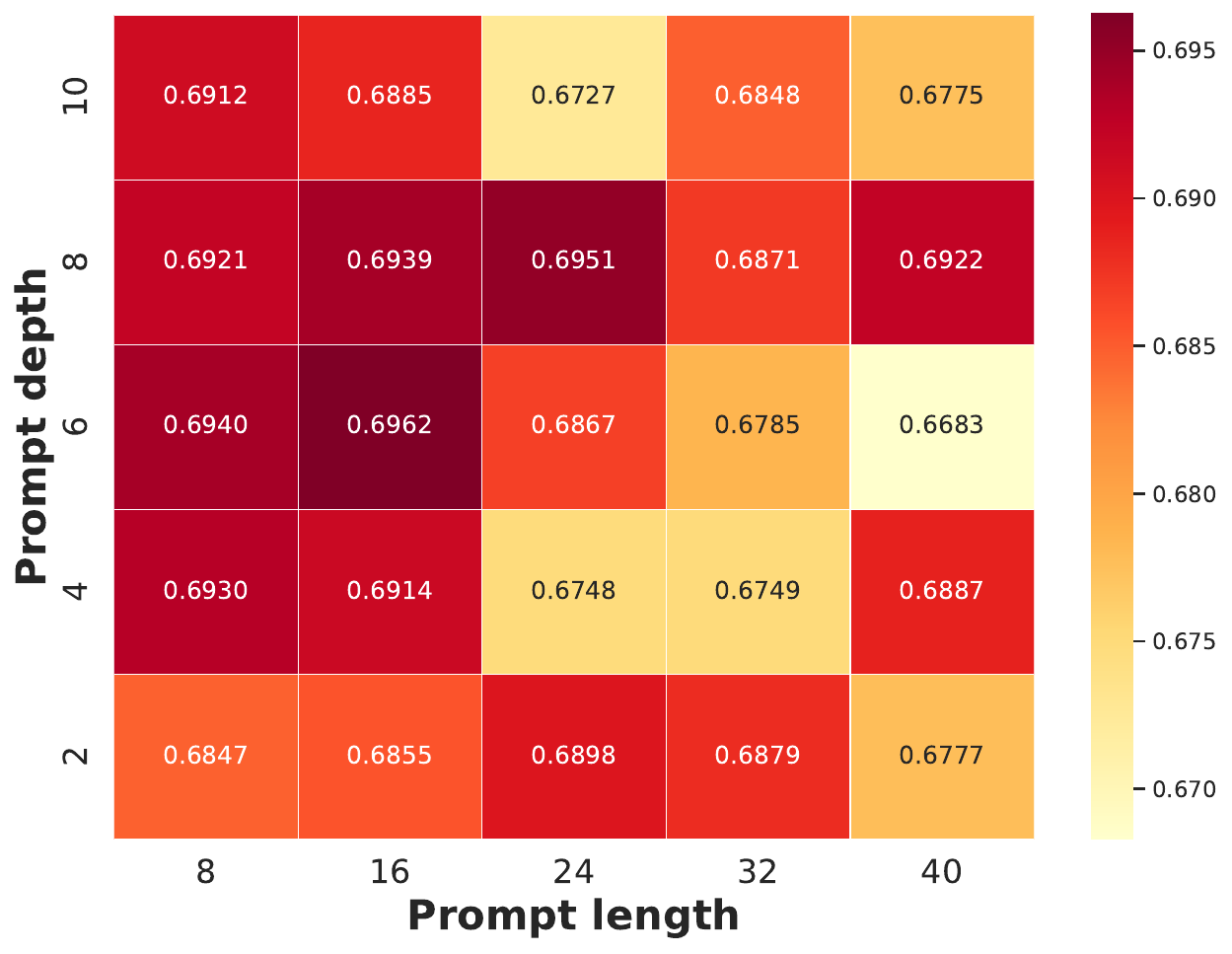}
    \caption{Ablation study on prompt length and depth on the Hateful Memes dataset~\cite{hatefulmemes} under the missing-both setting with a missing rate of $\eta = 70\%$. The best performance is achieved with a prompt depth of 6 and a prompt length of 16, indicating a balanced trade-off between representation capacity and optimization stability.} 
    \label{fig:absprompt}
\end{figure}

\subsection{Effect of Prompt Length and Depth.} 
\label{ab:sec4}

We conduct an ablation study on the prompt length and depth on the Hateful Memes dataset~\cite{hatefulmemes} under the missing-both setting with a missing rate of $\eta = 70\%$. As shown in Figure~\ref{fig:absprompt}, both factors have a noticeable impact on model performance. Increasing the prompt depth improves performance at shallow levels, while excessive depth yields diminishing returns. Similarly, longer prompts provide richer representations but may introduce redundancy when overly large. Overall, the best performance is achieved with a prompt depth of 6 and a prompt length of 16, suggesting that a moderate configuration strikes an effective balance between representation capacity and optimization efficiency.

\end{document}